# Two can play this Game: Visual Dialog with Discriminative Question Generation and Answering

Unnat Jain
UIUC
uj2@illinois.edu

Svetlana Lazebnik
UIUC
slazebni@illinois.edu

Alexander Schwing
UIUC
aschwing@illinois.edu

## Abstract

*Human conversation is a complex mechanism with subtle nuances. It is hence an ambitious goal to develop artificial intelligence agents that can participate fluently in a conversation. While we are still far from achieving this goal, recent progress in visual question answering, image captioning, and visual question generation shows that dialog systems may be realizable in the not too distant future. To this end, a novel dataset was introduced recently and encouraging results were demonstrated, particularly for question answering. In this paper, we demonstrate a simple symmetric discriminative baseline, that can be applied to both predicting an answer as well as predicting a question. We show that this method performs on par with the state of the art, even memory net based methods. In addition, for the first time on the visual dialog dataset, we assess the performance of a system asking questions, and demonstrate how visual dialog can be generated from discriminative question generation and question answering.*

## 1. Introduction

Human conversation is a complex mechanism with the intent to exchange information between at least two people. It is often very subtle and nuances are particularly important. It is hence an ambitious goal to develop artificial intelligence based agents for human-computer conversation about visual observations, that goes far beyond development of a simple question-answer mechanism.

Nonetheless, to obtain a basic understanding about how to construct artificial intelligence powered agents for conversation about visual observations, it is important to develop early prototypes using dialogues containing questions and answers. In a recent effort to facilitate this task, Das *et al*. [6] collected, curated and provided to the general public an impressive dataset which allows to design virtual assistants that can converse. Different from image captioning datasets, such as MSCOCO [21], or visual question answering datasets, such as VQA [2], the visual dialog dataset [6] contains short dialogues about a scene between two people. To direct the dialogue, the dataset was constructed by showing a caption to the first person ('questioner') to inquire more about the hidden image. The second person ('answerer') could see both the image and it's caption to provide answers to these questions.

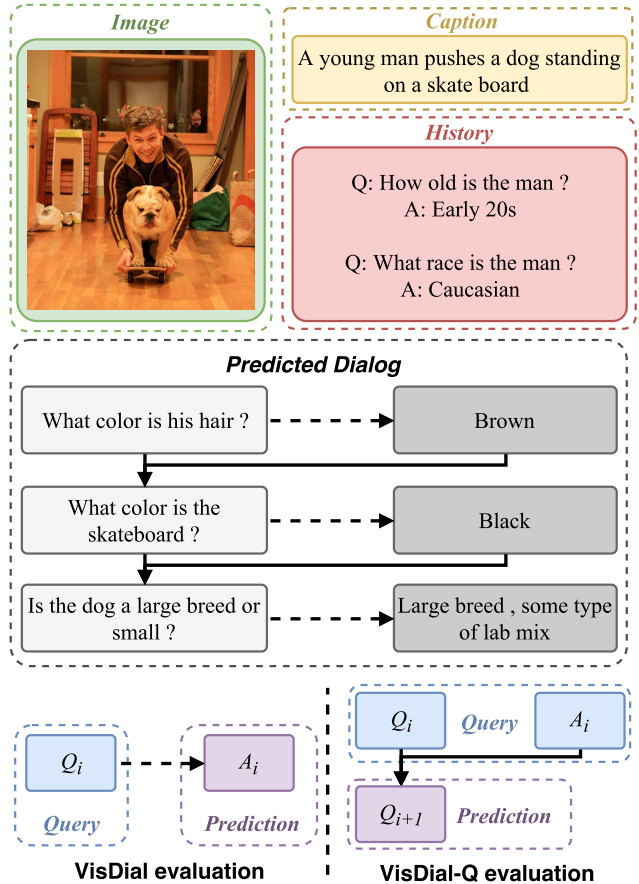

Figure 1: Visual dialog as a combination of two complementary tasks: (1) predicting a contextual answer to a given question (*VisDial* [6]); (2) predicting a contextual follow-up question to a given question-answer pair (*VisDial-Q*).

Beyond providing the Visual Dialog dataset (VisDial), to facilitate a fair comparison, Das *et al*. [6] suggest a concrete task that can be evaluated precisely. It requires the AI system to predict the next answer given the image, the question, and a history of question-answer pairs. A variety of discriminative and generative techniques were proposed, ranging from those based on Long-Short-Term-Memory (LSTM) units [12] to reasonably complex ones, which make use of memory nets [4] and hierarchical LSTM architectures.

In this paper we develop a deep net architecture that predicts an answer given a question, a caption, an image, and a question-answer history. The proposed approach outperforms existing baselines [6, 23] on the aforementioned answer prediction task. We present a careful assessment of its performance over five metrics. We also argue that the reverse setup, *i.e.*, prediction of the next question given the image, caption, and a history of question-answer pairs is equally important. We therefore re-purpose the visual dialog dataset and demonstrate that our developed architecture is applicable to this new question prediction setup without significant changes. In Fig. 1 we illustrate a combination of our models producing a visual dialog. To obtain this result, our discriminative questioning and answering modules communicate with each other.

## 2. Related Work

A conversation about an image or more generally an observation is hard to analyze, often very personal and even harder to predict. Despite or rather because of these difficulties, artificial intelligence based systems that master conversational capabilities are of great use, *e.g.*, for aiding visually impaired or for improving human-computer interaction.

Related to artificial intelligence agents that master conversation are several areas that have received a considerable amount of attention: (i) image captioning, *i.e.*, the task to describe the main content of an observed scene; (ii) visual question answering, *i.e.*, the task to answer a question about the content of a provided image; and (iii) visual question generation, *i.e.*, the task to generate a question about an observed scene. We briefly review each of those tasks in the following before discussing the visual dialog setup.

**Image Captioning:** Classical methods formulate image captioning as a retrieval problem. The best fitting description from a pool of possible captions is found by evaluating the fitness between available textual descriptions and images. This metric is learned from a set of available image descriptions. While this permits end-to-end training, matching image descriptors to a sufficiently large pool of captions is computationally expensive. In addition, constructing a database of captions that is sufficient for describing a reasonably large fraction of images seems prohibitive.

To address this issue, recurrent neural nets (RNNs) decompose the space of a caption into a product space of individual words. They have found widespread use for image captioning because they have been shown to produce remarkable results. For instance, [28] train an image CNN and a language RNN that shares a joint embedding layer. [41] jointly trains a CNN with a language RNN to generate sentences, [42] extends [41] with additional attention parameters and learns to identify salient objects for caption generation. [18] uses a bi-directional RNN along with a structured loss function in a shared vision-language space. Diversity was considered, *e.g.*, by Wang *et al*. [38].

**Visual Question Answering:** Beyond describing an image, a significant amount of research has been devoted to approaches which answer a question about a provided image. This task is often also used as a testbed for reasoning capabilities of deep nets. Using a variety of datasets [26, 33, 2, 9, 46, 17], models based on multi-modal representation and attention [24, 43, 1, 5, 8, 35, 40, 34], deep net architecture developments [3, 27, 25] and dynamic memory nets [39] have been discussed. Despite these efforts, it is hard to assess the reasoning capabilities of present day deep nets and differentiate them from memorization of training set statistics.

**Visual Question Generation:** In spirit similar to question answering but often involving a slightly more complex language part is the task of visual question generation. It has been proposed very recently and is still very much an open-ended topic. For instance, Ren *et al*. [33] discuss a rule-based algorithm which converts a given sentence into a corresponding question that has a single word answer. Mostafazadeh *et al*. [29] were the first to learn a question generation model using human-authored questions instead of machine-generated captions. They focus on creating a 'natural and engaging' question. Recently, Vijayakumar *et al*. [37] have shown preliminary results for this task as well. In contrast to the two aforementioned techniques, Jain *et al*. [15] argued for more diverse predictions and employed a variational auto-encoder approach. Work by Li *et al*. [20], introduce VQA and VQG as dual tasks and suggest a joint training for the two tasks. They leverage the state-of-the art VQA model by Ben-younes *et al*. [3] and achieve improvements for both VQA and VQG.

**Visual Dialog:** A combination of the three aforementioned tasks is visual dialog. Strictly speaking it involves both generation of questions and corresponding answers. However, in its original form [6], visual dialog required to predict the answer for a given question, a given image and a provided history of question-answer pairs. While this largely resembles the visual question answering task, a variety of different approaches have been proposed recently.

For instance, in [6], three models are formulated based on - late fusion, attention based hierarchical LSTM, and memory networks. A baseline for simple models is set using the 'late fusion' architecture. While late fusion has a simple architecture, the other two complex models obtain better performance. Following up, [23] proposed a generator-discriminator architecture where the outputs of the generator are improved using a perceptual loss from a pre-trained discriminator. The generator consists of an encoder (with two LSTM nets and attention mechanism) and a Gumbel-softmax [16] based LSTM decoder. The discriminator employs a similar encoder and a deep metric learning based loss. Unlike the methods of [6] which train in 4-8 epochs, [23] report pretraining of the generator and discriminator networks for 20 and 30 epochs respectively. Afterwards the generator is finetuned for additional epochs to obtain the final model.

Fig. 2 summarizes the difference between our approach

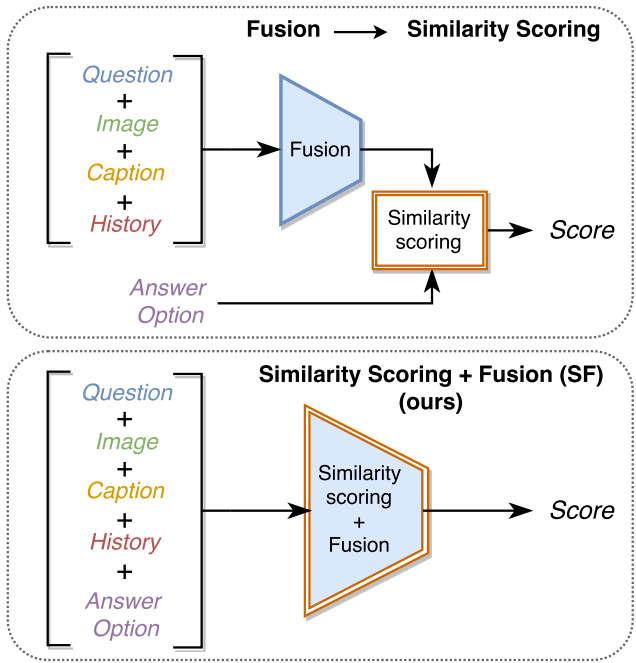

Figure 2: Overview of the proposed approach: Joint similarity scoring of answer option and fusion of all input features.

## 3. Approach

It is the purpose of this paper to maximally utilize the informativeness of options, *i.e.*, to use early option input. Hence, we focus on discriminative visual dialog systems. In contrast, generative methods model a complex output space distribution. Since discriminative frameworks cannot provide such free-form answers, they are restricted to environments where a small number of answers or questions is sufficient. Beyond focusing on the visual question answering part like [6], in this paper, we also provide results for question generation. We argue that this part is at least as important for a successful visual dialog system as answering a question.

To this end we develop a unified deep net architecture for both visual question answering and question generation. We will demonstrate in Sec. 4 that the proposed approach performs well on both tasks. In the following we first provide an overview of the proposed approach before we discuss the developed architecture in greater detail and provide implementation details. We finally discuss how we repurpose the visual dialog dataset to obtain a training set for the question generation task.

### 3.1. Overview

An overview of our approach is provided in Fig. 2. The visual dialog dataset contains tuples $(I, C, H_t, Q_t, A_t)$, consisting of an image $I$, a caption $C$, a question $Q_t$ asked at time $t \in \{1, \ldots, T\}$, its corresponding answer $A_t$, and a time dependent history $H_t$. $T$ is the maximally considered time horizon. The history itself is a set of past question-answer pairs, *i.e.*, $H = \{(Q_k, A_k)\}$ for $k \in \{1, \ldots, t-1\}$. At a high level, any visual dialog system, just like ours, operates on image embeddings, embeddings of the history and caption, and an embedding of the question. Generative techniques use embeddings of those three elements, or a combination thereof to model a probability distribution over all possible answers. Note that generative techniques typically don't take a set of answer options or their embeddings into account. In contrast, discriminative techniques operate on a set of answers, particularly their embeddings, and assess the fitness of every set member w.r.t. the remaining data, *i.e.*, the image $I$, the history $H_t$, the caption $C$ and the question $Q_t$. One member of the answer set constitutes the groundtruth, while other possible answers are assembled to obtain a reasonably challenging task.

### 3.2. Unified Deep Net Architecture

A detailed illustration of our architecture is provided in Fig. 3. Using LSTM nets we compute embeddings for the question at hand, the caption and the set of possible answer options. Similarly, to obtain an embedding for a question-answer pair, we use a question and an answer LSTM to encode all question-answer pairs in the history set $H$. Upon encoding the question and the answer of a question-answer and the existing methods for Visual Dialog. A study of similar type was done by Jabri *et al.* [14] for VQA. All aforementioned methods [6, 23] first encode the question, image, caption and history into a fused representation. Later this encoded representation is used to obtain similarity with the 100 answer options. In contrast, our model uses the answer option under evaluation as an early input. We perform both fusion and similarity scoring together using a multi-layer perceptron network. This joint optimization improves performance significantly compared to even memory networks [6]. We obtain quantitative results slightly better than the methods in [23]. Also, training of all our models converges within 5 epochs, which is significantly faster than the techniques proposed in [23].

Despite strong dialog information, the suggested evaluation of the VisDial dataset is strongly one-sided as mentioned before. To tackle this issue, Das *et al.* [7] introduced an image guessing game as a proxy to build visual question and answer bots. They adopt reinforcement learning based methods which they found to outperform maximum likelihood based supervised learning on respective metrics. Despite training both questioning and answering agents on the VisDial dataset, only answer metrics are reported. This is because at present there isn't an objective question generation protocol for the VisDial dataset. To bridge this gap, we provide a reconfiguration of the VisDial dataset, *i.e.*, 'VisDial-Q.' We introduce VisDial-Q to facilitate an evaluation of visual question generation agents. We also provide our baselines for VisDial-Q. We believe this reconfiguration to be useful for researchers that aim at evaluating the visual question generation side of the visual dialog task.

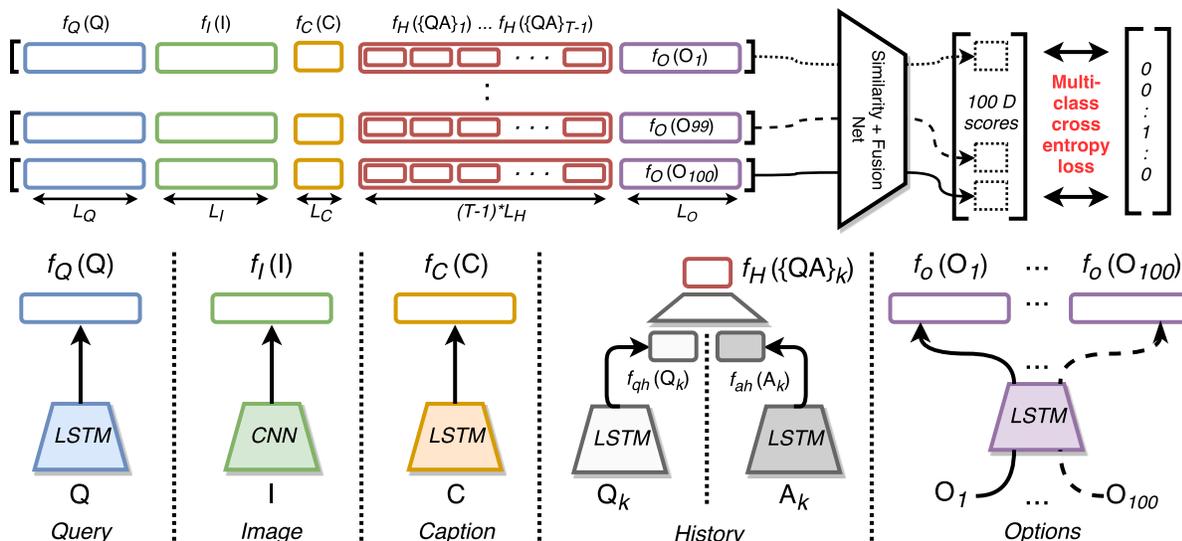

Figure 3: Architecture of our model for selecting the best answer option from a set of 100 candidates. LSTM nets transform all sequential inputs to a fixed size representation. The combined representations of $T-1$ previous question-answer pairs are concatenated to obtain the final history representation. Multi-class cross-entropy loss is computed by comparing a one-hot ground truth vector (based on the correct option) to output probabilities of the answer options.

pair in the history via the corresponding LSTM nets, we compute a single embedding by combining both representations via a fully connected layer. Concatenation of embeddings for all pairs in the history set $H$ constitutes the history embedding. We then concatenate the question embedding, the image embedding, the caption embedding, the history embedding, and the answer embedding for each of the possible answer options and employ a similarity network to predict a probability distribution over the possible answers. Since we score each option independently, our architecture works even if a different number of options are being evaluated at test time. We provide more details for each of the components in the following.

**Question and Answer Embedding:** The VisDial dataset questions are truncated to contain a maximum of $N_Q$ words. A $Stop$ token is introduced to mark the end of the question. Each word's $V$-dimensional one-hot encoding is transformed into a real valued word representation using a matrix $W_Q \in \mathbb{R}^{E_Q \times V}$. These $E_Q$-dimensional word embeddings are used as input for an LSTM which transforms them to $L_Q$-dimensional hidden state representations. The hidden state output corresponding to the last $Stop$ token is used as the sentence embedding of the question.

The methodology to obtain the representation for the answer options is identical. Each answer option is truncated to contain a maximum of $N_O$ words. $V$-dimensional one-hot representations of the words of an answer are transformed using a word embedding matrix $W_O \in \mathbb{R}^{E_O \times V}$. These $E_O$-dimensional word embeddings when transformed using an LSTM network give rise to an $L_O$-dimensional sentence embedding of the particular answer option at the last LSTM unit. If the question has 100 answer options, we extract a sentence embedding for each of the 100 options.

**Caption Embedding:** Similar to question and answer embeddings, captions are truncated to contain a maximum of $N_C$ words. Then a $Stop$ token is concatenated and these one-hot vectors are first transformed using an embedding matrix $W_C \in \mathbb{R}^{E_C \times V}$ before transformation into an $L_C$-dimensional caption embedding using an LSTM net $f_C(\cdot)$.

**Image Representation:** To obtain an image representation we make use of pretrained CNN features to represent images. For a fair comparison with baseline architectures proposed in [6], we use the activations of the second to last layer of the VGG-16 [36] deep net. We normalize these $L_I$-dimensional activations by dividing via their $\ell_2$ norm, as also performed in [6].

**History Embedding:** All question-answer pairs $(Q_k, A_k)$ before the query time $t$, $i.e.$, $k \in \{1, \ldots, t-1\}$, serve as history. An embedding matrix $W_{qh} \in \mathbb{R}^{E_{qh} \times V}$ maps one-hot word vectors to real valued embeddings. These are transformed by a question-history LSTM $f_{qh}(\cdot)$ to an $L_{qh}$-dimensional sentence embedding. Similarly, the answer-history is encoded via $W_{qh} \in \mathbb{R}^{E_{qh} \times V}$ and $f_{ah}(\cdot)$ to obtain an $L_{ah}$-dimensional sentence embedding. Both the question and answer embedding are combined using a fully connected layer to obtain an $L_H$-dimensional representation of a pair $\{(Q_k, A_k)\}$. The number of question-answer pairs before the current query is variable ($t-1 \in [0, T-1]$). Existing models tackle this issue of variable length history in different ways. For instance, Das $et$ $al$.'s [6] 'Late Fusion' (LF) concatenates words of all previous questions and answers and transforms it using another LSTM network. They also implement a hierarchical LSTM to address this challenge. Their model performing best in terms of accuracy is based on a memory network which maintains every previous question and answer as a 'fact entry.' Lu $et$ $al$. [23] use an attention based mechanism to combine all previous rounds of history to get a single representation. On the con-trary, we found a very simple method to be effective. We introduce an $Empty$ token to our vocabulary of words (which already includes the stop token $Stop$). For all the missing question-answer rounds, we pass the $[Empty, Stop]$ sequence to the $f_{qh}(\cdot)$ and $f_{ah}(\cdot)$ LSTM nets. Using this we always have $T-1$ embeddings of question-answer pairs. A concatenation results in the $(T-1) \cdot L_H$-dimensional history representation.

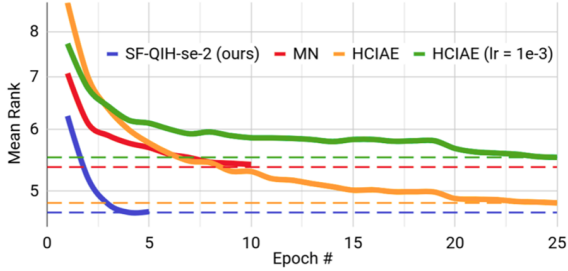

Figure 4: Comparison of our method to state-of-the-art discriminative models–memory networks (MN) [6] and HCIAE [23]. We use the authors' implementations. HCIAE-D-NP-ATT is the best performing discriminative model proposed by [23], which we abbreviate as HCIAE.

**Similarity Scoring + Fusion Network:** The individual representations of the question, image, caption, history as well as an answer option are concatenated to form an ensemble. This ensemble is represented by an $L_S = L_Q + L_I + L_C + (T-1)*L_H + L_O$ dimensional vector. As mentioned before, unlike previous methods, we perform similarity scoring and feature fusion jointly. This is achieved using a multi-layer perceptron (MLP). To reduce the number of parameters, the MLP is structured to have a decreasing number of activations for the intermediate layers before arriving at a single scalar score for each $L_S$-dimensional representation. During inference we choose the answer option having the highest score. During learning the answer option scores are transformed into probabilities using a softmax layer. We report results of architectures with MLP having one and two hidden layers. The single hidden layer MLP has $\lfloor L_s/2 \rfloor$ hidden nodes. The two hidden layered MLP has $\lfloor L_s/2 \rfloor$ and $\lfloor L_s/4 \rfloor$ nodes in its intermediate representation layers.

To simplify training, we employ Batch Normalization [13] layers after every linear layer which we found to be more robust. We normalize before the ReLU non-linearity, as suggested in [13].

### 3.3. Network Training

To describe training more formally, let $F_w(O_i)$ denote the score for answer option $i$ obtained from the 'similarity scoring + fusion network,' and let $w$ denote all the parameters of the architecture illustrated in Fig. 3. For simplicity we avoid to explicitly mention other inputs such as the query, the image, *etc*. While inference chooses the highest scoring answer $i^* = \arg\max_i F_w(O_i)$ given learned parameters $w$, training optimizes for the parameters $w$ via the multi-class cross-entropy loss:

$$\min_w \sum_{\mathcal{D}} \left( \ln \sum_{\hat{i}=1}^{100} \exp F_w(O_{\hat{i}}) - F_w(O_{i^*}) \right),$$

where $\mathcal{D}$ denotes the dataset containing ground truth information $i^*$. All our models are trained using the Adam optimizer [19] with a learning rate of $10^{-3}$.

We experimented with both normal initialization by He *et al*. [11] and Xavier normal initialization [10] and found the former to work better in our case for both MLP and LSTM weights. We found that sharing the weights of the language embedding matrices greatly helps in learning better word representations. Two hidden layered MLP nets assessing similarity and fusing the representations consistently performed better than a one layered MLP. We use the data splits suggested in [6] for VisDial v0.9: 80k images for training, 3k image for validation and 40k for test. We use the validation set to determine when training doesn't progress any further and report metrics on the test set. All our models converge in under 5 epochs of training on this dataset, which is illustrated in Fig. 4

### 3.4. Implementation Details

The VisDial dataset has ten rounds of question-answer pairs, hence $T = 10$. $N_Q$, $N_A$ and $N_C$ are set to 20, 20 and 40 respectively. Dimensions of all embeddings, *i.e.*, $E_Q, E_O, E_C, E_{qh}$ and $E_{ah}$ are set to 128. LSTM hidden state dimension of query and options, *i.e.*, $L_Q$ and $L_O$, are set to 512. LSTM hidden state dimension of caption, question-history and answer-history, *i.e.*, $L_C, L_{qh}$ and $L_{ah}$, are set to 128. All the LSTMs are single layered. In accordance to the baselines of [6], we use pretrained VGG-16 relu7 features for the image embedding, hence, $L_I = 4096$. Note that on the contrary, [23] utilize 25k dimensional VGG-19 pool5 features. Also, [23] report their result after making use of 82k training images which is more than the 80k images suggested in [6] for VisDial v0.9. Finally, [23] utilize deep metric learning and a self-attention mechanism to train a discriminator network which leverages the availability of answer options. We achieve this via a simple LSTM-MLP approach. However, it must be noted that [23] also investigate generative models for the VisDial dataset which we don't explore here.

### 3.5. VisDial-Q Dataset and Evaluation

Das *et al*. [6] highlight the challenge of evaluating dialog systems and they propose to evaluate individual responses at each round of the dialog. To this end they create a multiple choice retrieval setup as a 'VisDial evaluation protocol.' As explained earlier, the system is required to choose one out of 100 answer options for a given question. The image, caption and previous question-answer pairs can be leveraged by the system to help make this choice. However, no surrogate task for assessment of question generation is provided.

| Model | MRR | R@1 | R@5 | R@10 | Mean |
|---|---|---|---|---|---|
| *Query only* | | | | | |
| LF-Q [6] | 0.5508 | 41.24 | 70.45 | 79.83 | 7.08 |
| SF-Q-1 | 0.5619 | 42.11 | 72.12 | 81.39 | 6.55 |
| SF-Q-se-1 | 0.5651 | 42.32 | 72.54 | 81.83 | 6.39 |
| SF-Q-se-2 | **0.5664** | **42.45** | **72.75** | **81.98** | **6.32** |
| *Query + Image only* | | | | | |
| LF-QI [6] | 0.5759 | 43.33 | 74.27 | 83.68 | 5.87 |
| SF-QI-1 | 0.5940 | 45.49 | 75.95 | 85.19 | 5.40 |
| SF-QI-se-1 | 0.5964 | 45.72 | 76.25 | 85.64 | 5.29 |
| SF-QI-se-2 | **0.6010** | **46.19** | **76.73** | **85.95** | **5.18** |
| *Query + Image + Caption + History* | | | | | |
| LF-QIH [6] | 0.5807 | 43.82 | 74.68 | 84.07 | 5.78 |
| HRE-QIH [6] | 0.5868 | 44.82 | 74.81 | 84.36 | 5.66 |
| MN-QIH [6] | 0.5965 | 45.55 | 76.22 | 85.37 | 5.46 |
| HCIAE [23] | 0.6222 | 48.48 | 78.75 | 87.59 | 4.81 |
| SF-QIH-1 | 0.6101 | 47.04 | 77.69 | 86.78 | 5.00 |
| SF-QIH-se-1 | 0.6207 | 48.19 | 78.66 | 87.53 | 4.79 |
| SF-QIH-se-2 | **0.6242** | **48.55** | **78.96** | **87.75** | **4.70** |

Table 1: *VisDial* evaluation metrics. '-1' and '-2' denote one and two hidden MLP layers respectively. '-se' denotes shared embedding matrices for all LSTMs.

To test the questioner side of visual dialog, we therefore create a similar '*VisDial-Q evaluation protocol*.' A visual question generation system is required to choose one out of 100 next question candidates for a given question-answer pair. To do this it may utilize the image, caption and previous question-answer pairs. What is left to answer is how these 100 candidates for the next question are selected.

We closely follow the methodology adopted by Das *et al.* [6] to select 100 answer candidates from the visual dialog dataset of the human question-answer rounds. We select 100 candidate follow-up questions to a given question-answer (QA) pair as the union of the following four sets:

**Correct**: The next question asked by the human is the ground truth question.

**Plausible**: Plausible questions are follow-up questions to the 50 most similar QA pairs in the dataset. Similar QA pairs are found by comparing concatenated GloVe embeddings [30] of the QA pair being considered with the representation of other QA pairs. Question GloVe embeddings are obtained following [6], *i.e.*, (1) concatenate the GloVe embedding of the first three words of the question; (2) average the GloVe embeddings of the remaining words; and (3) concatenate both vectors. Answer GloVe embeddings are obtained by averaging the GloVe embeddings of all its words. $\ell_2$ distance computed on the concatenated question and answer GloVe embeddings is used to find nearest neighbor QA pairs. We make sure that a nearest neighbor QA pair is not from the same image (same as [6]). Additionally, for the *VisDial-Q* evaluation, we also ensure that the nearest neighbor QA pair isn't the last ($10^{th}$) QA round of a dialog, as no human follow-up question is available.

**Popular**: Question possibilities also contain the 30 most popular questions of the original dataset.

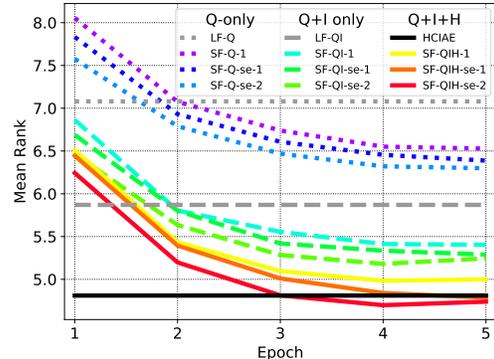

Figure 5: *VisDial* evaluation: Mean rank values for our models and best models from [6, 23]

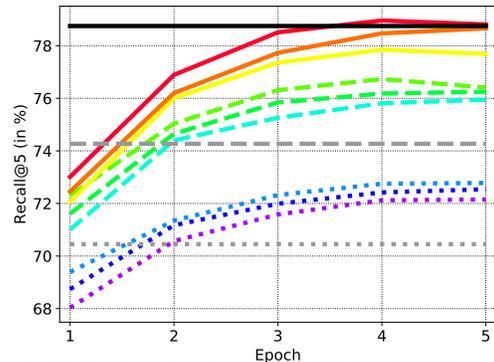

Figure 6: *VisDial* evaluation: Recall@5 values for our models and best models from [6, 23] (same legend as Fig. 5)

**Random**: The remaining question options which are left to complete a set of 100 unique candidates are filled with random questions from the dataset.

Our intention for creating a set of question options using this methodology is analogous to [6]. These candidates encourage an algorithm to distinguish between correct, plausible, and popular candidates.

At this point it is important to address a strong difference in the nature of evaluating a module for generating an answer from a technique producing a question. While answering a given question based on options (and some additional information) has fairly little randomness, the questioning analog is significantly more challenging. That is, for a given QA pair, there could be more than one 'correct' follow-up question in the options. Despite this inherent ambiguity, objective evaluation of the question generation procedure is equally important. It depicts the questioning system's ability to rank a human generated question. The system should be encourage to rank the human generated question in its top ranks, if not at the highest one. Therefore, the ensemble of metrics proposed in [6] and described in Sec. 4.2 is even more important than a single Recall@1 based evaluation.

Our deep net architecture developed for the answering task in Sec. 3.2 can be deployed for the *VisDial-Q* task, with almost no adjustments. Since there exists no follow-up question to the last QA pair in a dialog of the VisDial dataset, the maximally considered time horizon $T$ is 9 for the VisDial-Q dataset. The 'query' for the original visual dialog task is a question whose answer we wish to choose. On the other hand, 'query' for the questioning side of visual dialog (VisDial-Q) is a QA pair for which we wish to choose the most relevant follow-up question. For VisDial-Q evaluation, words of the QA pair (concatenation of question and answer words) serve as input to the 'query' LSTM in Fig. 3. The options $O_1, \ldots, O_{100}$ are now candidate follow-up questions, instead of candidate response answers. All other parameters are identical to the ones mentioned in Sec. 3.3 and Sec. 3.4.

| Model | MRR | R@1 | R@5 | R@10 | Mean |
|---|---|---|---|---|---|
| *Query only* | | | | | |
| SF-Q-1 | 0.1909 | 9.18 | 26.18 | 38.87 | 23.03 |
| SF-Q-se-1 | 0.1936 | 9.57 | 26.20 | 38.66 | 22.99 |
| SF-Q-se-2 | **0.1950** | **9.70** | **26.44** | 38.67 | **22.92** |
| *Query + Image only* | | | | | |
| SF-QI-1 | 0.2953 | 16.82 | 41.58 | 56.27 | 14.57 |
| SF-QI-se-1 | 0.2970 | 17.06 | 41.60 | 56.05 | 14.48 |
| SF-QI-se-2 | **0.3021** | **17.38** | **42.32** | **57.16** | **14.03** |
| *Query + Image + Caption + History* | | | | | |
| SF-QIH-1 | 0.3877 | 25.03 | 53.03 | 68.33 | 10.09 |
| SF-QIH-se-1 | 0.4028 | 26.51 | 54.74 | 69.95 | 9.54 |
| SF-QIH-se-2 | **0.4060** | **26.76** | **55.17** | **70.39** | **9.32** |

Table 2: *VisDial-Q* evaluation metrics. '-1' and '-2' denote one and two hidden layers in MLP respectively. '-se' denotes shared embedding matrices for all LSTMs.

## 4. Experiments

In the following we evaluate our proposed architecture on prediction of both answers and questions. To this end, we first provide details about the datasets and evaluation metrics used. We then discuss our quantitative assessment before providing qualitative results.

### 4.1. Datasets

We train our models on the VisDial v0.9 dataset [6] which currently contains over 123k image-caption-dialog tuples. Each dialog has 10 question-answer pairs. The images are unique and are obtained from the MSCOCO [21] train and validation split. The dataset was collected by recording a conversation between two people on Amazon Mechanical Turk. The first person is only provided the caption to start the conversation, and is tasked to ask questions about the hidden image to better understand the scene. The second person has access to both image and caption and is asked to answer the first person's questions. Both are encouraged to talk in a natural manner, which is markedly different from [2]. Due to this setup, the obtained question-answer pairs have inherent temporal continuity and are also visually grounded. The VisDial v0.9 train, validation and test sets consists of 80k, 3k and 40k images.

### 4.2. Evaluation Metrics

Many popular metrics like BLEU, ROUGE and METEOR are empirically shown to have low correlation with human judgement of dialog systems [22]. For an objective evaluation of visual dialog systems, [6] suggests metrics for predicted rank of the correct answer option. These are Recall@1, Recall@5, Recall@10, Mean Reciprocal Rank, and Mean Rank of the ground truth answer. Recall@$k$ is the percentage of questions for which the correct answer option is ranked in the top $k$ predictions of a model. Mean Rank is the empirical average of the rank allotted by a model to the ground truth answer option. Mean Reciprocal Rank is the empirical average of 1/rank allotted by a model to the ground truth answer option. Lower values for Mean Rank and higher values for all the other metrics are desirable.

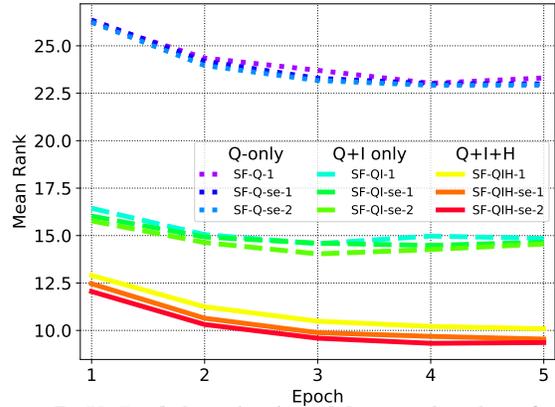

Figure 7: *VisDial-Q* evaluation: Mean rank values for our models.

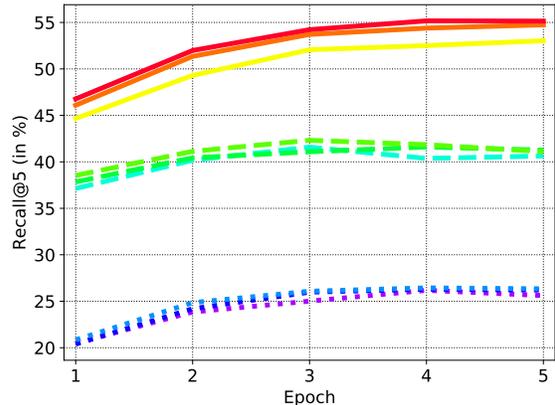

Figure 8: *VisDial-Q* evaluation: Recall@5 values for our models. (same legend as Fig. 7)

### 4.3. Quantitative Assessment

In the following we provide a quantitative assessment of our approach. We first discuss results for the question answering task before focusing on question generation.

**Visual Question Answering:** The performance of the proposed architecture for predicting a contextual answer to a given question (VisDial evaluation) is presented in Tab. 1. We gradually increase context from only question (Q), to question and image (QI), and finally all given context (QIH). Our 'similarity scoring + fusion' (SF) performs best in all three scenarios. Adding image and history cues improves results. We provide the metrics for baselines from existing work evaluating on the VisDial dataset. This includes models proposed in [6], based on late fusion (LF), hierarchical LSTM net (HRE), and memory networks (MN). Another important baseline is the best performing discriminative model (HCIAE-D-NP-ATT) [23]. We use the abbreviation HCIAE for this model. Fig. 5 and Fig. 6 compare the mean rank and recall@5 of different models. Our SF-QIH model achieves 78.96% recall@5 and 4.70 mean rank.

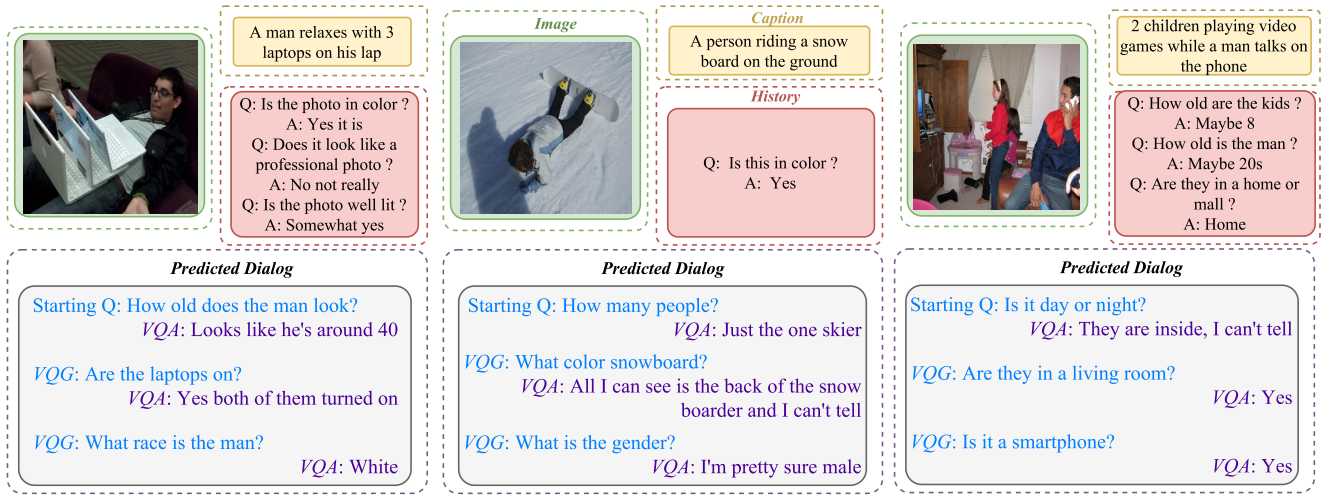

Figure 9: Joint unrolling of questioning and answering modules on test images. The VQG module chooses the most relevant next question based on previous QA pairs and context.

**Visual Question Generation:** A similar evaluation of the proposed architecture for the task of predicting the next question based on a given QA pair and context (VisDial-Q evaluation) is presented in Tab. 2. By closely investigating our results, we obtain some intuitive insights. First, without any context, predicting the next question is a much more difficult task than answering a question without context. This can be observed from the average mean rank for VisDial-Q ($\sim$ 20) in comparison to the average mean rank for VisDial ($\sim$ 7). Second, large improvements when comparing Q vs QI and QI vs QIH suggest that image and history cues are much more important for the question prediction task than for answer prediction. Figs. 7, 8 compare the mean rank and recall@5 of different models. Our SF-QIH model achieves a 55.17% recall@5 and 9.32 mean rank.

### 4.4. Qualitative Evaluation

In this section we discuss qualitative results. Instead of presenting two separate qualitative evaluations of our architecture on the answering and questioning side of visual dialog, we provide a joint analysis. After completing the answering task of choosing the best option for a given question, we provide this QA pair to our pretrained question generation module. The newly generated question is then again put up for discriminative answering by the answering module. Hence we 'generate' dialog using our discriminative models. Fig. 9 summarizes a few of those unrolled examples. A few arrangements are necessary to jointly unroll our discriminative questioning and answering modules, since answer options and next question options are available for only dataset dialogs, while we are 'generating' (i.e., selecting) new sequences. Hence we need to create options on the fly, by choosing from a set of questions and answers of nearest neighbor images. We uniformly sample one of the top 10 ranking questions chosen by the question module to add some more diversity. We again emphasize that these dialogs are 'generated' by choosing from a set of options, which differs from truly generative approaches.

Based on the observed empirical results we conclude that our models capture cues from all three contexts, image, caption and history. There are questions pertaining to partially visible objects, which can be attributed to the caption cue. The same is true for objects visible in the images which aren't mentioned in the history/caption text. We experimented with different number of rounds of initial history - 1, 2, 3 and 5. In all cases, our models choose relevant follow-up questions and fairly correct answers. Since there are no groundtruth options for these predicted dialog sequences, we can't report quantitative metrics for this dynamic setup where our models communicate with each other.

## 5. Conclusion

We developed a discriminative method for the visual dialog task, i.e., predicting an answer given question and context. Our approach outperforms existing baselines which often use complex architectures. More importantly, our approach can be applied with almost no change to prediction of a question given context, which we think is equally important. We introduce the VisDial-Q evaluation protocol to quantitatively assess this task and also illustrate how to combine both discriminative methods to obtain a system for visual dialog. Going forward we plan to combine visual dialog and textual grounding [31, 32, 45, 44].

**Acknowledgements:** We thank NVIDIA for providing the GPUs used for this research. This material is based upon work supported in part by the National Science Foundation under Grants No. 1563727 and 1718221, and Samsung.

## Supplementary Material

The supplementary document is organized as follows:

- Sec. 6 covers additional details of VisDial-Q.
- Sec. 7 shows additional quantitative evaluations beyond Mean Rank and Recal@5 (included in the main paper).
- Sec. 8 shows additional qualitative examples of unrolling question generation and answering modules.

## 6. VisDial-Q and VisDial comparison

Sec. 3.5 explains the re-purposing of VisDial to VisDial-Q, using correct, plausible, popular and random question options. Here we include a comparison of the distribution of answers and questions. Sentence distribution of target questions are shown in Fig. 10. A steeper slope of answer distribution vs. question distribution shows the challenging nature of question generation. This is supported by entropy (higher 4.71bits for question and 4.52bits for answer). Fig. 10b has examples of popular question candidates.

## 7. Quantitative Results

In the following we present additional quantitative results, some of which were already mentioned in the paper. We report Recall@1, Recall@5, Recall@10, Mean Reciprocal Rank (MRR) and Mean rank for the test sets of both answer prediction task (*VisDial evaluation*) and question prediction task (*VisDial-Q evaluation*). Fig. 11 and Fig. 12 summarize these metrics as training proceeds for the two tasks. Our models perform significantly better than the most complex architectures of [6]. Our models are easy to train, with convergence in under 5 epochs in contrast to a 20-30 epoch pre-training required for the baseline set by generator-discriminator architecture in [23]. Since we introduce a new evaluation protocol for question prediction in visual dialog, there aren't any existing baselines for this task in Fig. 12.

## 8. Qualitative Results

As mentioned in the paper, we decide to unroll both our question prediction and answer prediction module together to show how these discriminative models can be used to 'generate' dialog. The answer module chooses the best answer option to a given question while the question module chooses the best next question to a given question-answer pair. As mentioned in the paper, a few arrangements are necessary to jointly unroll questioning and answering modules, since answer options and next question options are available for only dataset dialogs, while we are 'generating' (*i.e.*, selecting) new sequences. We create options on the fly, by choosing from a set of questions and answers of nearest neighbor images. Since there are no ground-truth options for these predicted dialog sequences, we can't report quantitative metrics for this dynamic setup where our models communicate with each other.

We test our models in two different visual dialog setups. *Firstly*, we unroll our VQA and VQG modules when there is very little history. A visual dialog system needs to be more inquisitive in such a setup and '*explore*' the image. Fig. 13 shows both short and long dialogs predicted by our models in such a setup. *Secondly*, we also test our models when there is a long history available to build on. Here, the models need to be consistent with existing context - avoid repetitions, and handle co-reference resolution. In such a setup the models '*exploit*' the available history to find finer details about the image. The generations do not repeat questions from the history and reference objects using correct pronouns. Fig. 14 shows both short and long visual dialogs predicted by our discriminative VQA and VQG models.

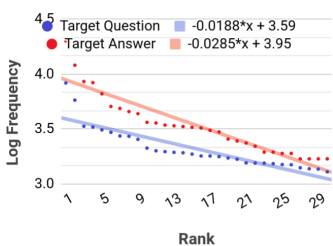

(a) Comparing target answer and target question distributions

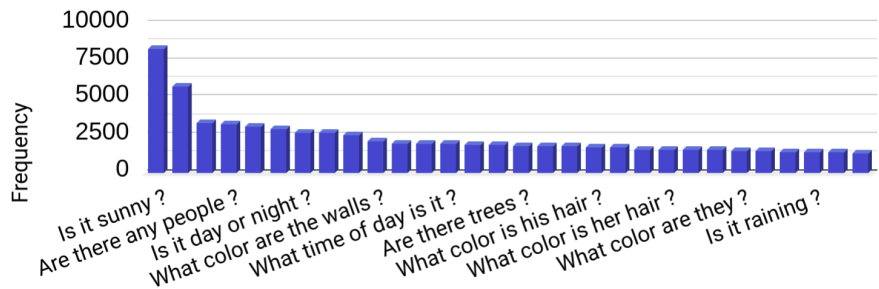

(b) Target question distribution (top 30)

Figure 10: (a) compares target distribution of questions and answers (top 30 ranked targets). Steeper slope of answers indicates higher frequency biases in the answer targets. (b) displays the frequency distribution of questions, analogous to Fig. 15 in [6].

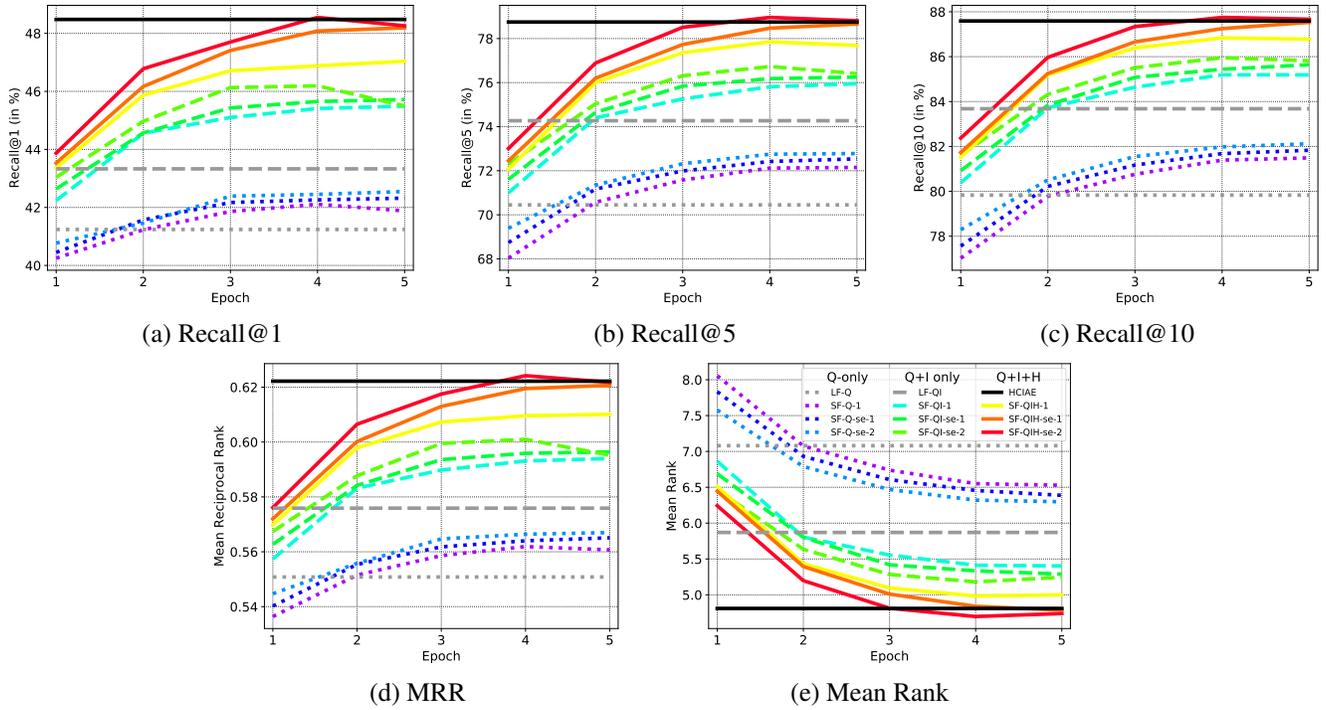

Figure 11: *VisDial* evaluation protocol: Evaluation metrics for our models and best models from [6, 23] - Late fusion (LF) and HCIAE-D-NP-ATT (abbreviated as HCIAE). '-1' and '-2' refer to one and two hidden layers in our 'similarity learning + fusion net' (SF) model. '-se' refers to shared word embeddings across all LSTM nets. (Legend is same as (e))

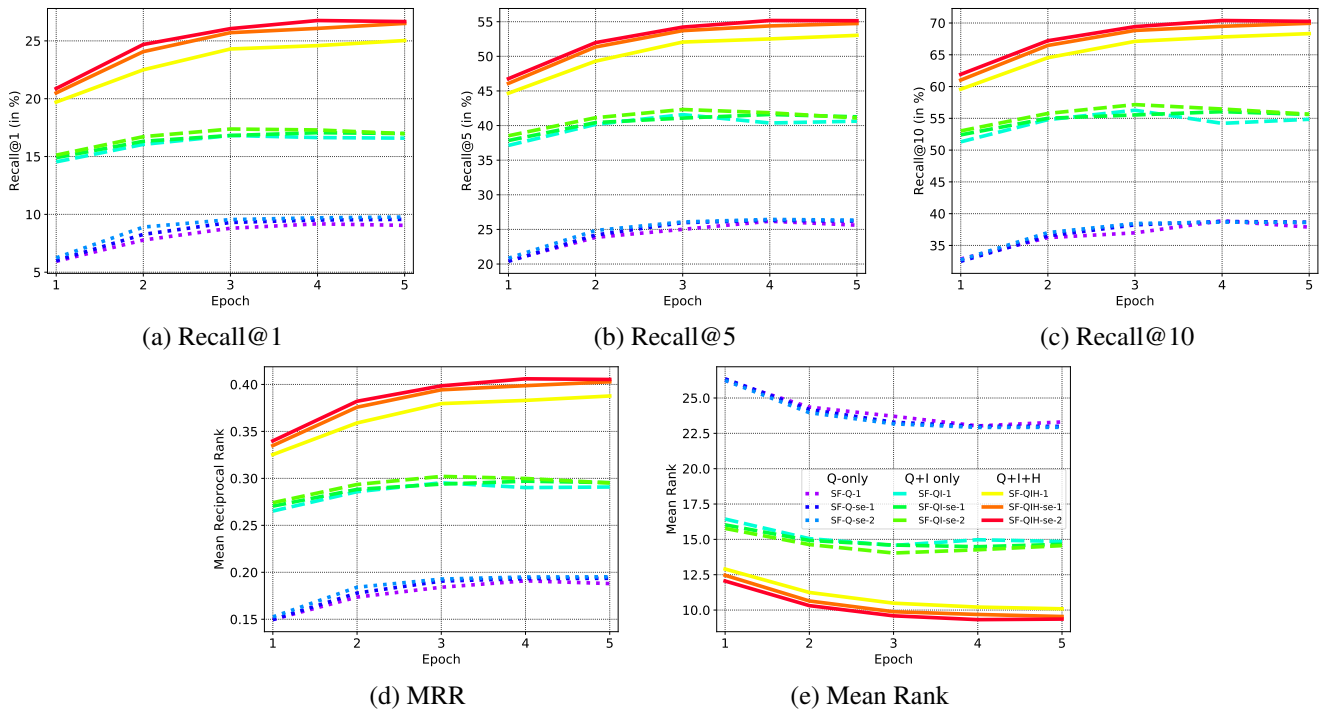

Figure 12: *VisDial-Q* evaluation protocol: Metrics for our models on the newly proposed VisDial-Q evaluation protocol. '-1' and '-2' refer to one and two hidden layers in our 'similarity learning + fusion net' (SF) model. '-se' refers to shared word embeddings across all LSTM nets.

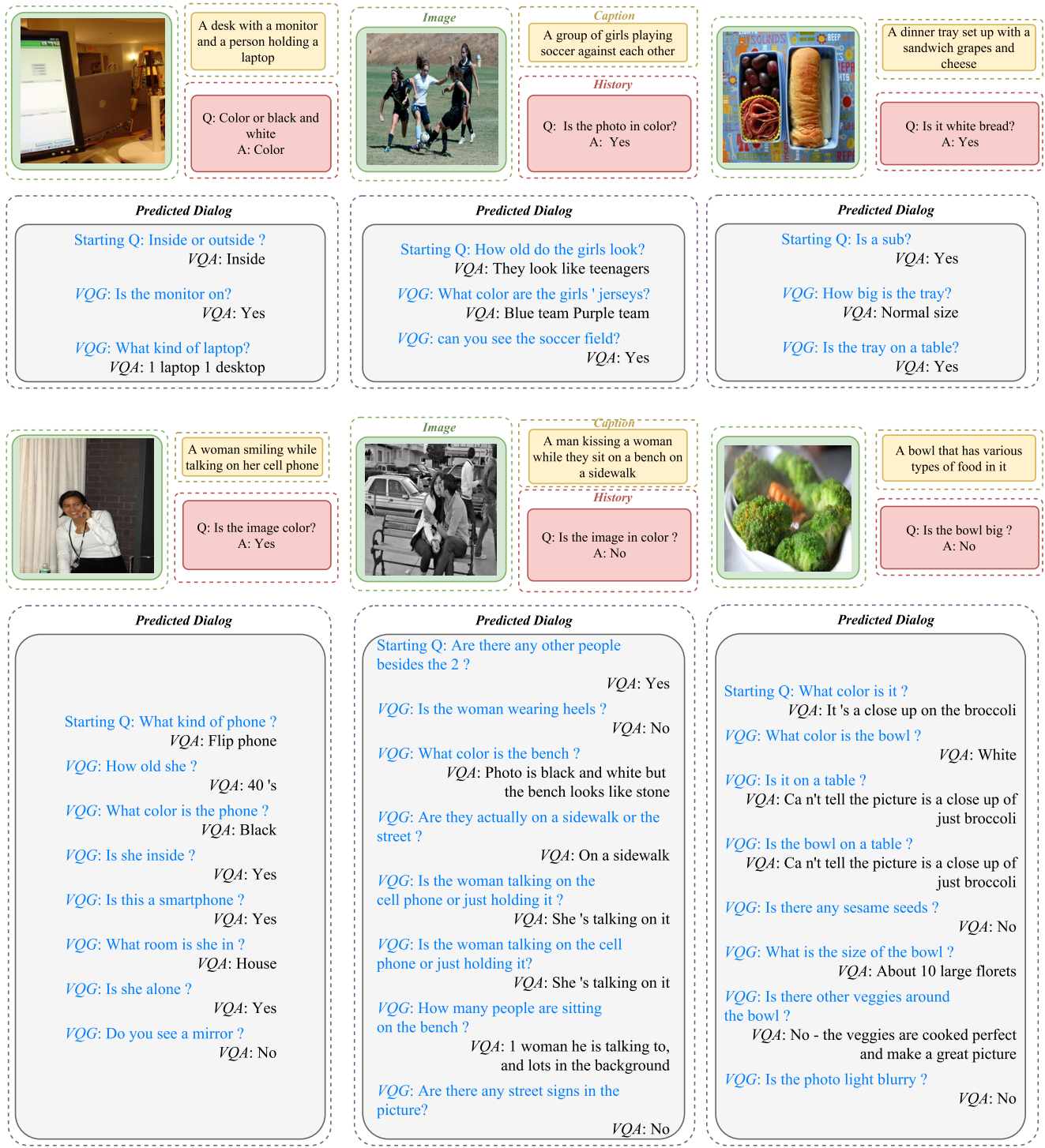

Figure 13: Joint unrolling of VQA and VQG modules for short history (1 QA pair): Short and long dialogs 'generated' by our discriminative models.

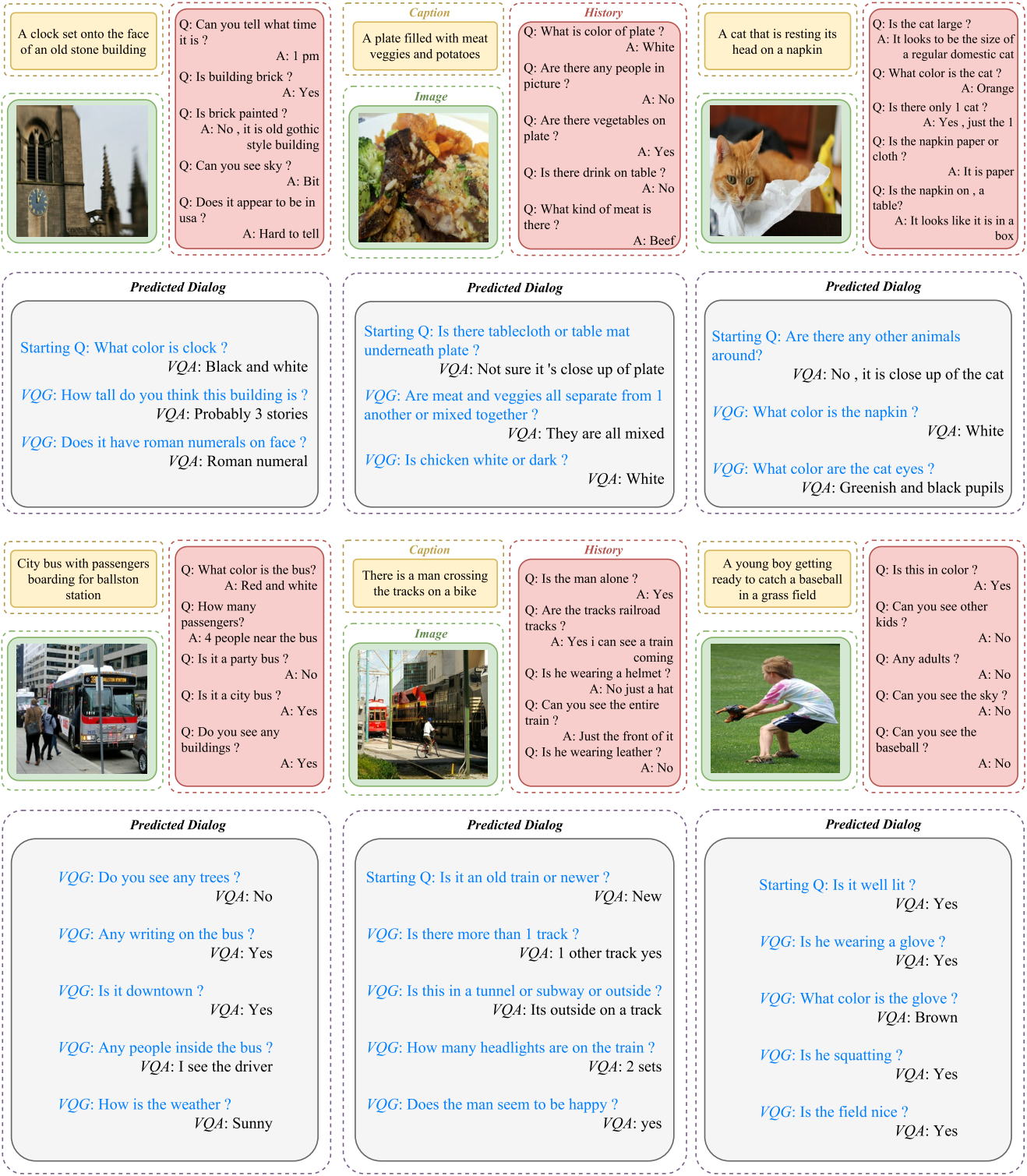

Figure 14: Joint unrolling of VQA and VQG modules for long history (5 QA pair): Short and long dialogs 'generated' by our discriminative models.